\pdfoutput=1

\documentclass[11pt]{article}

\usepackage[final]{acl}

\usepackage{times}
\usepackage{latexsym}

\usepackage[T1]{fontenc}

\usepackage[utf8]{inputenc}

\usepackage{microtype}

\usepackage{inconsolata}

\usepackage{graphicx}
\usepackage{subcaption}
\usepackage{caption}

\usepackage{amsmath}
\usepackage{enumitem}
\usepackage{xcolor}
\usepackage{dblfloatfix}

\title{Columbo: Expanding Abbreviated Column Names for Tabular Data\\[1mm] Using Large Language Models}

\author{\textbf{Ting Cai, Stephen Sheen, AnHai Doan}\\
  University of Wisconsin-Madison\\
  \texttt{\{tcai35, sheen2, ahdoan\}@wisc.edu}}

\begin{document}
\maketitle
\begin{abstract}
  Expanding the abbreviated column names of tables, such as ``esal''
  to ``employee salary'', is critical for many downstream NLP
  tasks for tabular data, such as NL2SQL, table QA, and keyword
  search. This problem arises in enterprises, domain sciences,
  government agencies, and more. In this paper, we make three
  contributions that significantly advance the state of the
  art. First, we show that the synthetic public data used by prior work
  has major limitations, and we introduce four new datasets in
  enterprise/science domains, with real-world abbreviations. Second,
  we show that accuracy measures used by prior work seriously
  undercount correct expansions, and we propose new synonym-aware
  measures that capture accuracy much more accurately.  Finally, we
  develop Columbo, a powerful LLM-based solution that exploits
  context, rules, chain-of-thought reasoning, and token-level
  analysis. Extensive experiments show that Columbo significantly
  outperforms NameGuess, the current most advanced solution, by
  4-29\%, over five datasets. Columbo has been used in production on EDI,
  a major data lake for environmental sciences.
\end{abstract}

\section{Introduction}\label{intro}

Tabular data is ubiquitous in companies, domain sciences, government
agencies, and others \cite{nameguess,edi}. Using this data well,
however, has been difficult. A major reason is that the names of the
tables and columns are often abbreviated, appearing quite cryptic (see
Figure \ref{ex1} and Appendix \ref{sample-cols}).  This makes it hard
for downstream NLP tasks to process the tables.

In particular, a recent work \cite{nameguess} shows that using abbreviated
column names significantly reduce accuracy for three NLP tasks,
NL2SQL, schema-based relation detection, and table QA, by 10.54\%,
40.5\%, and 3.83\%, respectively. That work also argues that expanding
column names improves the readability of the tables, enables data
integration, and improves keyword search (to discover relevant
tables). Another recent work \cite{snails} shows that abbreviated
column names significantly reduce the accuracy of NL2SQL. Our own
experience working with a major data lake (edirepository.org) also shows that expanding column names
improves the accuracy of keyword search, schema matching (i.e.,
finding columns that are semantically the same), and column annotation
with the concepts in a given ontology (see ``Columbo in Production''
in the experiment section).

Thus, table/column name expansion is a core challenge for the fast
growing direction of NLP tasks for tabular data, and has received
growing attention \cite{nameguess,snails,sawant2024nlp,anonymous2025realistic,singh2025leveraging}.
The goal is to expand table/column names
into meaningful English phrases, such as ``eSal'' to ``Employee
Salary'', ``eDTPh'' into ``Employee Day Time Phone'',
``1997lgNutSExt'' into ``1997 Long-term Nutrient Study Experiment'',
and so on. Clearly, this can enormously help downstream
applications. For example, given a user query ``employee phone'', a
keyword search application can correctly return the table
``EMPS(eName, eSal, eDTPh, ...)'' if it knows that ``eDTPh'' means
``Employee Day Time Phone''.

\begin{figure}[t]
  {\footnotesize
\begin{verbatim}
EMPS(eName, eSal, eDTPh, ...)
Ref_Sectors_GICS(GICS_IND_GRP_CD, ...)
1997lgNutSExt(Date, canWt, hclWt, corWetWt, ...)
\end{verbatim}
}
\caption{Examples of abbreviated table and column names in companies
  and domain sciences.}\label{ex1}
\end{figure}

As far as we can tell, the most advanced work for this challenge is
NameGuess in EMNLP-2023 by Amazon AWS \cite{nameguess}. That work focuses on
column name expansion and frames it as a natural language generation
problem.  It proposes a solution that uses LLMs to obtain 69.3\%
exact-match accuracy (using GPT4), in contrast to 43.4\% accuracy
obtained by human. NameGuess has clearly showed the promise of
using LLMs to expand abbreviated column names. But it has three major
limitations, as we discuss below. In this paper we describe Columbo,
which addresses these limitations and significantly advances the
state of the art.

First, NameGuess experimented with just one dataset, which is
\emph{public data} from the cities of San Francisco, Chicago, and Los
Angeles (a.k.a. Open City Data). We show later that LLMs achieve lower
accuracy on \emph{enterprise data} (coming from companies) and
\emph{domain science data}. Thus, we believe that \emph{a good
benchmark for column name expansion cannot contain just public
data}. In this work we introduce four more datasets that come from
enterprises and domain sciences.

Another problem is that NameGuess \emph{synthetically} creates the
abbreviated column names, e.g., by randomly dropping, shuffling, or
replacing characters from English phrases. So many abbreviated column
names in its dataset look ``unrealistic'', e.g., ``r'' for ``area'',
``mmj'' for ``medical'', and LLMs struggle to correctly expand such
names. Given that we do not yet have good methods to synthetically
create abbreviated column names, we believe that \emph{a good
benchmark for column name expansion should also contain abbreviated
column names that come from real data}.  The four datasets introduced
in this paper contain column names abbreviated by humans. Later we
show that LLMs indeed can leverage their vast knowledge about
real-world abbreviation patterns to correctly expand these names.

The second limitation is that NameGuess computes accuracy
in a restricted way. The most important accuracy measure,
exact match, computes the fraction of columns where the name predicted
by LLMs \emph{exactly matches} the ``gold'' name. This penalizes cases
of minor variations, e.g., ``geography identifier'' vs ``geographical identifier'',
``photo credit'' vs ``picture credit'', etc. To solve this, we 
introduce a new accuracy measure called ``synonym-aware exact match''.
We show that this new measure captures the performance of column name
expansion solutions much more accurately.

The third limitation is that NameGuess uses a rather basic LLM
solution. It simply asks the LLM to expand the names of the 
columns (given a few expansion examples). We have developed a
significantly more powerful solution. Our solution provides a lot of
context information to the LLM, e.g., the name of the target table and
the topics of similar tables (we infer these topics using LLMs). We ask the LLM to follow a set of rules
(that help generate the correct expansions) and use chain-of-thought
reasoning.  Finally, we reason about column name expansion at
\emph{the token level}, i.e., we translate each column name into a
sequence of tokens, expand each token into an English phrase, then
combine the phrases to obtain the column name expansion. Together,
these features help our solution significantly improve accuracy
compared to NameGuess. In summary, we make the following contributions:
\begin{itemize}
\item We show that synthetic public data is not sufficient for
  evaluating solutions for column name expansion. We introduce four new
  non-public datasets with real-world abbreviated column names.

\item We show that computing accuracy via exact string matching is problematic.
  We introduce a new measure that captures accuracy much
  more accurately.

\item We develop Columbo, an LLM-based solution that exploits
  context, rules, chain-of-thought reasoning, and token-level analysis.

\item We provide extensive experiments that show that Columbo outperforms
  NameGuess on all five datasets, improving the absolute accuracy by 4-29\%,
  and the relative accuracy by 4-46\%.
\end{itemize}  
Columbo has been used in production on EDI, a major data lake
for environmental sciences. We briefly describe this experience
in Section \ref{exper}. The code and datasets (except Finance
and University, for which we do not have permission to release) are
available at github.com/anhaidgroup/columbo.

\section{Problem Definition}\label{probdef}

Similar to NameGuess, given a set of tables (e.g., those in a data lake), we seek to expand
the column names. Expanding the table names is more complicated,
as we discuss in Section \ref{exper}, and hence is deferred to future work.

We assume that each column name $c$ can be represented as a sequence
$t_1d_1\ldots d_{n-1}t_n$, where each $d_i$ is a \emph{delimiter} (i.e., a
special character such as `\_', `-', the space character, or the empty
character) and each $t_i$ is a \emph{token} that can be expanded into a
meaningful English phrase $e(t_i)$. Phrase $e(t_i)$ must contain all
characters of token $t_i$, in that order.

For example, column name ``eSal'' can be tokenized into tokens ``e''
and ``Sal'', separated by the empty-character delimiter. Token ``e''
expands to ``Employee'' and token ``Sal'' expands to ``Salary''. Other
expansion examples are ``Rm'' $\rightarrow$ ``Room'' and
``CD''$\rightarrow$ ``Certificate Deposit''. The expansion $e(c)$ of
column $c$ is then the concatenation of the expansions of its tokens.

In many application contexts, we cannot access the data tuples of
the tables, for reasons of privacy, compliance, performance, etc.
\cite{semtab}. So here we consider the input to be just the
(abbreviated) table names and column names.

NameGuess shows that state-of-the-art hosted LLMs such as GPT-4
achieves the highest accuracy for column name expansion \cite{nameguess}. As a result,
here we focus on these LLMs, specifically on GPT-4o. In future work we
will consider open-source LLMs that can be deployed in-house.

\section{New Datasets}\label{sec:dataset_creation}\label{datasets}

\begin{table*}[t]
    \centering
    \includegraphics[width=0.9\linewidth]{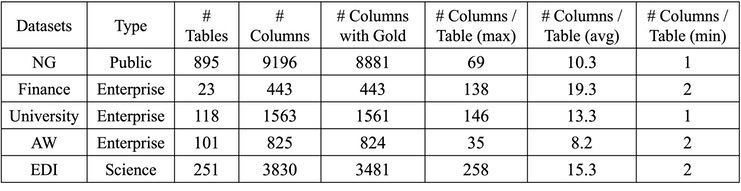}
    \caption{Statistics of the 5 datasets used in our experiments.}
    \label{data}
\end{table*}

We now make the case that a good benchmark for column name expansion
cannot contain just public data with synthetic abbreviations. It must
also contain non-public data with real-world abbreviations. We then
introduce four new such datasets.

Specifically, NameGuess uses just one dataset obtained from the Open
Data Portals of San Francisco, Chicago, and Los Angeles, covering
business, education, health, etc. To evaluate expansion solutions,
NameGuess needs both abbreviated column names and the gold (i.e.,
correct) expansions. To do this, NameGuess uses heuristic rules to find
column names that are meaningful English phrases. It abbreviates
these names, e.g., by randomly shuffling, dropping, or replacing
characters. Finally, it applies solutions to the abbreviated names to
see if they can recover the original column names.

Our experiments with the NameGuess dataset revealed two
problems. First, we found that LLMs achieve a much higher accuracy on
this dataset compared to the four enterprise/science datasets (e.g.,
81.5\% vs 63.2-73.8\% EM accuracy, see Table \ref{overall}). Prior
work has observed the same phenomenon for other data tasks
\cite{stone}, presumably because today LLMs have been trained on a lot
more public data than enterprise and science data. Put differently, we
speculate that the NameGuess dataset covers popular concepts (e.g.,
transportation, education, etc.) that are ubiquitous online, whereas
the remaining four datasets cover ``rarer'' concepts (e.g., specific
to a vertical). Thus, if we want to apply expansion solutions to
non-public data, we cannot rely on experiments with just public data,
as the results can be misleading.

To address this problem, in this paper we use the five datasets described in
Table \ref{data}. ``NG'' is the NameGuess dataset. ``Finance'' and ``University''
are two datasets obtained from companies, covering the finance and academic domains.
``AW'' is a variation of the AdventureWork dataset released by Microsoft, and ``EDI''
is a dataset from the environmental science domain. Appendix \ref{adatasets}
discusses how we generated these datasets. 

The second problem with the NameGuess dataset is that it is difficult
to accurately mimic human's abbreviation patterns. Thus, many
synthetic abbreviated column names look ``unrealistic'', and LLMs
struggle to expand these. Table \ref{overall} shows that on the NameGuess
dataset, LLMs can only improve EM accuracy from 81.5\% to 85.2\%.  In
contrast, the four new datasets have real-world column names already
abbreviated by their human creators. Here Table \ref{overall} shows that LLMs
can exploit their vast knowledge about real-world abbreviation
patterns to achieve high accuracy, improving EM accuracies from
63.2-73.8\% to 87.6-93.3\%. We conclude that a good benchmark for this
problem should also contain real-world abbreviated column names, as our
four new datasets do.

\section{New Accuracy Measures} \label{sec:synonym_metric}\label{acc}

NameGuess uses three accuracy measures. EM computes the fraction of
columns where the predicted expansion $x$ exactly matches the gold
expansion $g$. Word-level $F_1$ is $2PR/(P+R)$, where $P$ is the
fraction of tokens in $x$ that occur in $g$, and $R$ is the fraction
of tokens in $g$ that occur in $x$. BERT-score $F_1$ is computed
similarly, except that two tokens are considered equivalent if the
cosine similarity score of their BERT-based embedding vectors is high
\cite{nameguess}.

\begin{table}[t]
    \centering
    \includegraphics[width=1\linewidth]{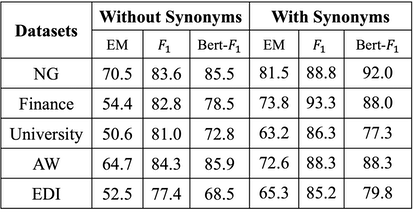}
    \caption{NameGuess accuracies without and with synonyms.}
    \label{tsyns}
\end{table}

\begin{figure*}[t]
    \centering
    \includegraphics[width=1\linewidth]{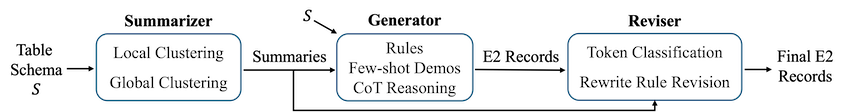}
    \caption{The overall architecture of Columbo.}
    \label{arch}
\end{figure*}

Among these three measures, EM is most intuitive, but as defined, it is
too restrictive. It regards cases of minor variations, e.g.,
``geography location'' vs ``geographical location'', ``picture
credit'' vs ``photo credit'', as not matched. To address this problem,
we examined the five datasets described in Section \ref{datasets}, and created
synonym pairs, e.g., ``geography'' = ``geographical'', ``picture'' =
``photo''. Given a gold expansion $g$, we use these synonym pairs to
create all gold variations, e.g., creating ``geography location'' from
``geographical location''. Then we declare that a predicted expansion
$x$ matches $g$ if it matches any variation of $g$. We call this new
measure \emph{synonym-aware EM}. We modify word-level $F_1$ and BERT-score
$F_1$ similarly to be synonym aware.

One may wonder if the BERT-based $F_1$ score captures ideas similar to
the synonym-based EM approach. We found that this is not the case,
because there are many synonyms that are very specific to a particular
vertical. Being trained mostly on public data, BERT is not aware of
such synonyms. Examples include "time 0" vs "initial", "site" vs
"hub", "log" vs "logarithm", "met" vs "meteorological" in EDI, "prime"
vs "principal" in University, and "sedol" vs "stock exchange daily
official list" in Finance.

Table \ref{tsyns} shows the accuracies of applying the NameGuess
solution to all five datasets. Clearly, the accuracy measures without
synonyms significantly underestimate the true accuracies. For example,
they show 50.6-70.5\% EM, whereas the measures with synonyms show
63.2-81.5\% EM. Furthermore, the relative ranking of dataset
difficulty as measured by the two EM measures does change. Consider
Finance and AW. Table \ref{tsyns} shows that the EM measure with no
synonyms ranks Finance harder than AW (54.4\% vs 64.7\%). But the
synonym-aware EM measure ranks Finance easier than AW (73.8\% vs
72.6\%). For the rest of this paper, we use the synonym-aware
accuracy measures.

\section{The Columbo Solution}\label{method}

We now describe the Columbo solution, which improves
upon NameGuess.  Given a table $T$, NameGuess sends batches of 10
column names from $T$ to the LLM.  It provides several examples of
column name expansion, e.g., ``c\_name''$\rightarrow$ ``customer
name'', then asks the LLM to expand the above 10 column
names. NameGuess does not exploit any additional information, e.g.,
the table name.  In contrast, Columbo exploits the names of table $T$
and related tables, rules, chain-of-thought reasoning, token-level
analysis, and more.

Specifically, the key insight behind Columbo is that, to solve this
problem well, we should use all available information and reason at a
deeper level. Consequently, we use LLMs because to expand a column
name correctly, we need a lot of domain knowledge. LLMs have been
trained on tons of data and can be viewed as very large stores of
domain knowledge. So applying LLMs to this problem is promising.
Second, we observed that when expanding column names, LLMs keep making
similar mistakes. This is why we formulate a set of general rules (in
the prompt) telling LLMs not to make these mistakes. Third, we
observed that there is a well-known chain-of-thought (CoT) process
that a human typically follows to expand a column name: first expand
each token, then combine these token-level expansions to obtain the
column-level expansion. This suggests that CoT can be well matched to
this problem. Finally, we perform token-level analysis by identifying
tokens with potentially incorrect expansion rules and fixing those.

Figure \ref{arch} describes the Columbo architecture, which consists
of 3 modules: Summarizer, Generator, and Reviser. We now describe these
modules, then discuss the rationales behind the design decisions. 

\begin{figure}[t]
    \centering
    \includegraphics[width=1\linewidth]{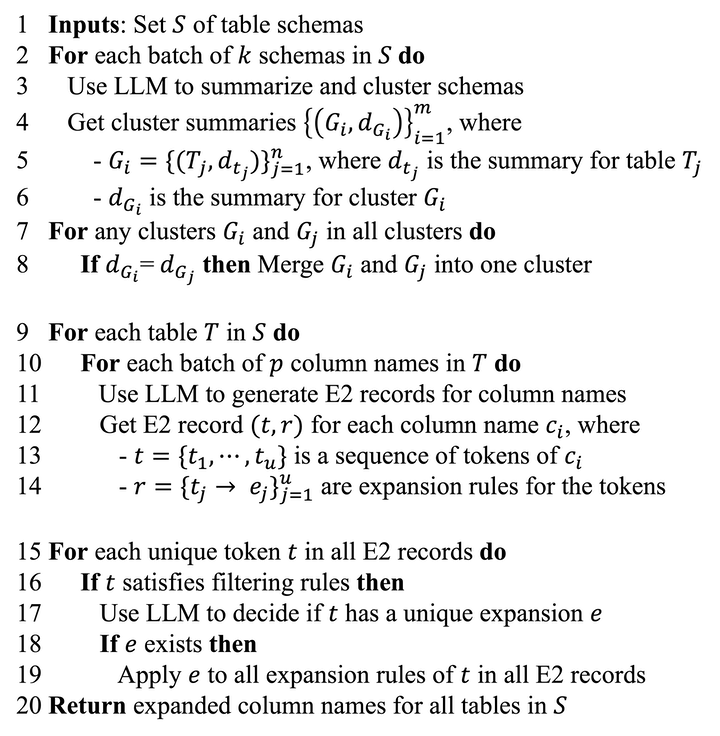}
    \caption{The pseudo code of Columbo.}
    \label{pseudo}
\end{figure}

\paragraph{The Summarizer:} This module creates two kinds of summaries
to be used by subsequent modules (see Lines 1-8 of
the algorithm in Figure \ref{pseudo}). Let $S$ be the set of table schemas for which we want
to expand the column names. We first send batches of $k$ table schemas
from $S$ to the LLM, and ask it to cluster the $k$ tables into groups
(currently $k = 30$). For each group $G$, we ask the LLM to provide a
\underline{group summary} $d_G$, which is a short English phrase that best summarizes
$G$, and similarly, for each table $T$ in group $G$, we also ask for a
\underline{table summary} $d_T$ that best summarizes $T$.

Appendix \ref{asum} shows the prompt for a sample batch and the output of the
LLM. For example, it shows that 3 tables BusinessEntity(ModDate,
bID,rID), BusinessEntityAddress(ModDate, aID, aTypeID, bID, rID),
BusinessEntityContact(ModDate, PersonID, bID, cTypeID, rId) have been
clustered into a group, with group summary ``Business Entity
Structure'', and that Table BusinessEntity(ModDate, bID,rID) has the
summary ``Represents a generic business entity that can be a person,
vendor, or customer''.  

After processing all tables in $S$ (via batches of up to 
$k$ tables), we perform a ``global merge'' that merges all
groups with the same summary. Thus, the Summarizer produces a
clustering of all input tables into groups, where each group $G$ has a
summary $d_G$, and each table $T$ has a summary $d_T$.

\paragraph{The Generator:} This module expands the column names (see Lines 9-14 of
the algorithm in Figure \ref{pseudo}). Specifically, for each table $T$ (in the set $S$), we
send batches of $p$ columns of $T$ to the LLM, and ask it to expand
the $p$ column names (currently $p = 10$). We structure the LLM prompt
for each batch as follows.

First, we provide the context, which is the name of table $T$, as well
as \underline{the names and table summaries} of up to $q$ tables (randomly sampled) from the same group as $T$
(currently $q = 100$).

Second, we specify a set of rules, e.g., ``expand all abbreviations in a column name'', ``do not
expand numbers'', ``do not add extra words or explanations'', etc.

Finally, unlike NameGuess which just asks the LLM for the expansion $e(c)$ of a given column name $c$,
we ask the LLM to provide a chain-of-thought reasoning that leads to the expansion. Specifically,
we ask the LLM to parse column name $c$ into a sequence of tokens, then provide the expansion
for each token, then concatenate these expansions into $e(c)$. We provide a few examples of such
reasoning in the prompt. 

Thus, for each input column name $c$, this module produces
\underline{a sequence of tokens} $t_1\ldots t_n$, and
\underline{expansion rules for the tokens}: $t_1\rightarrow e_1,
\ldots, t_n\rightarrow e_n$.  We call this output \underline{the E2
  record} for column $c$ (where ``E2'' stands for ``expansion \&
explanation''). The expansion of $c$ is then the string
$e_1\ \ldots\ e_n$. Appendix \ref{agen} shows the prompt for a sample
set of columns and the output of the LLM.

\paragraph{The Reviser:} This module improves upon the output of the Generator, by identifying
tokens with potentially incorrect expansion rules, then trying to fix those (see Lines 15-20 of
the algorithm in Figure \ref{pseudo}). 

Specifically, we first process the E2 records (output by the
Generator) to identify the set $P$ of all tokens that have more than
one expansion rule, e.g., $dt\rightarrow date$ and $dt\rightarrow
data$.  Next, for each token $x\in P$, we ask the LLM if $x$ should
have just one expansion rule, and if so, to identify that
rule.

To help the LLM make the above decisions, we provide it with \underline{the
summaries of all groups}, and all expansion rules of token $x$ (that we have
identified from the E2 records). For each expansion rule of $x$, we
also provide its frequency (i.e., the number of column names in which
that expansion rule is used), and a sample table schema in which that expansion
rule is used.

We optimize the above process by using a set of rules to
decide which tokens in $P$ to send to the LLM. For example, if a token
$x\in P$ has too few characters (currently set to 1), then we do not
send $x$ to the LLM, because it is likely that $x$ has more
than one correct expansion, e.g., $e\rightarrow employee$ and
$e\rightarrow electronic$.

Suppose the LLM has identified a set $Q$ of pairs $(x,e)$, where token
$x$ should always be expanded into $e$. Then we use $Q$ to modify the
E2 records (produced by the Generator). The Reviser outputs the
modified E2 records as final records, and the expansions of the column
names can be quickly obtained from these E2 records. Appendix
\ref{arev} shows the prompt for a sample token and the output of the
LLM. 

\paragraph{Discussion:} We now discuss the rationales behind the major
design decisions. Consider the Generator. This module asks the LLM to
perform chain-of-thought reasoning in which it translates the column
name into a sequence of tokens, then finds the expansion of each
token. We found that this improves the LLM's accuracy. Further, this
gives us the tokens and their expansion rules, which enable
token-level analysis, such as the one carried out by the Reviser. We
also found that just supplying examples of column name expansion was
not enough. LLMs were still prone to producing incorrect output, e.g.,
adding extra words, explanations, changing word orders, etc.  Adding
rules asking the LLM not to do so helps improve accuracy (see the
experiments).

Intuitively, providing context can help expand column names. For
example, the LLM may incorrectly expand column RUSS\_CD to ``Russian
Code''.  But being told this column is in Table RUSSELL\_INDEX, it can
correctly expand the column to ``Russell Code''. So we provide the LLM
with the name of the target table $T$. We also cluster all the input
tables so that we can find the tables related to $T$, and provide a
subset of these tables to the LLM, as additional context. To provide
this subset of tables, we can just send their schemas to the LLM. But
it turns out that many of these tables can have a large number of columns
(e.g., 200+), making their schemas too large. This is why in the
Summarizer we ask the LLM to provide for each table a short table
summary. Then instead of sending the full table schema, we just send
the (shorter) table name and summary.

Now consider the Reviser. If we can provide the LLM with
information about the entire dataset, i.e., the entire set of table
schemas $S$, it can more accurately decide if a token $x$ has just
one expansion in $S$ (e.g., RUSS should always be expanded to ``Russell'').
But sending all table schemas in $S$ to the LLM is impractical. So
in the Summarizer we ask the LLM to provide a short group summary
for each group of tables, then send the LLM just the summaries
of all groups. 

\section{Experiments}\label{sec:experiments}\label{exper}

We now evaluate Columbo, using the five datasets described in Table
\ref{data} (see Section \ref{datasets}). We conducted all experiments
using GPT-4o, version gpt-4o-2024-08-06, with temperature 0, max
completion tokens 6000, and default values for all other parameters.

\begin{table}[t]
    \centering
    \includegraphics[width=0.98\linewidth]{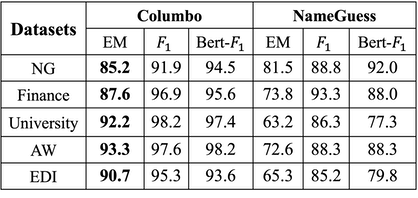}
    \caption{The accuracy of Columbo vs. NameGuess}
    \label{overall}
\end{table}

\paragraph{Overall Performance:} Table \ref{overall} compares Columbo with NameGuess, the state-of-the-art
solution, using the three synonym-aware accuracy measures described in
Section \ref{acc}. In what follows we focus on the EM accuracy measure, as it
is most intuitive. First, the table shows that Columbo significantly
outperforms NameGuess on all five datasets, improving the absolute EM
accuracy by 4-29\% and the relative EM accuracy by 4-46\%.

Second, NameGuess achieves lower EM accuracies on the enterprise
(Finance, University, AW) and science (EDI) datasets, compared to the
public (NG) dataset: 63.2-73.8\% vs 81.5\%. This suggests that LLMs
perform worse on non-public data (prior work \cite{stone} has reached
the same conclusion).

\begin{table}[t]
    \centering
    \includegraphics[width=0.9\linewidth]{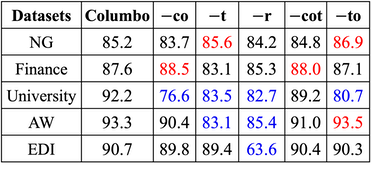}
    \caption{Ablation studies for Columbo.}
    \label{ablation}
\end{table}

Finally, Columbo is able to improve the EM accuracy of NameGuess on
the non-public datasets by a large amount, from 63.2-73.8\% to
87.6-93.3\%, but it ``struggles'' to improve the EM accuracy of
NameGuess on the public dataset NG, from 81.5\% to just 85.2\%. We
believe this is because many column names in NG look ``unrealistic'',
as they are synthetically abbreviated (as we discussed in Section \ref{acc}).
So the LLM fails to expand many such columns, resulting
in a small EM improvement. Overall, the results in Table \ref{overall}
suggests that a good benchmark for column name expansion should
contain non-public data with real-world abbreviated column names. 

\paragraph{Ablation Studies:} We now evaluate the major components of Columbo.
Table \ref{ablation} shows the EM accuracies of Columbo (the 2nd column)
and the five Columbo versions in which we remove a major component. (Appendix \ref{aablation}
shows the full result which also contains the remaining two accuracy measures.)

First we modify Columbo to not exploit any context information, i.e.,
removing the Summarizer and disabling using the table and group summaries
in the Generator and Reviser. The EM accuracies of this Columbo version are
reported in Column ``-co'' in Table \ref{ablation}.

Second, we modify Columbo to not exploit table names. Specifically, we
need table names in the Summarizer to create the summaries, so we keep
the Summarizer as is.  But we remove all mentions of table names in
the Generator and Reviser, using only table summaries where
appropriate. The results are in Column ``-t''.

Third, we remove all nine rules used in the Generator (see Column
``-r''). Fourth, we modify the Generator to not use chain-of-thought
reasoning. The results are in Column ``-cot''. Finally, we remove the
token-level analysis by disabling the Reviser, and report the results
in Column ``-to''.

Table \ref{ablation} shows that disabling a component typically leads to a
drop in accuracy, sometimes by a lot, as highlighted in blue in the table, e.g., 92.2\% of Columbo vs 76.6\% of ``-co''
on University, 90.7\% of Columbo vs 63.6\% of ``-r'' on EDI. Occasionally the
accuracy increases, as highlighted in red in the table. 
But this increase is minimal, from 0.2-1.7\%. Thus, the results suggest
that the components contribute meaningfully to the overall accuracy of Columbo.

\begin{table}[t]
\centering
\includegraphics[width=0.9\linewidth]{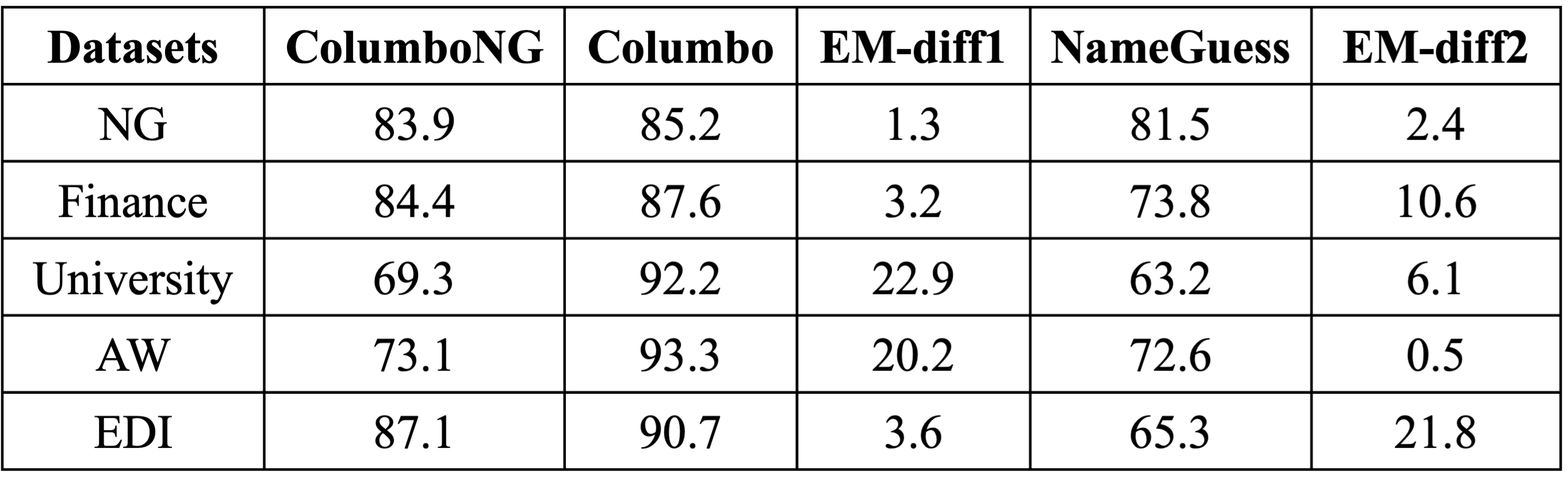}
\caption{Exact Match accuracy of ColumboNG vs. NameGuess}
\label{colng_ng}
\end{table}

We also examined the case where Columbo uses exactly the same input
information as NameGuess. To do so, we develop ColumboNG, by removing
the table clustering step and token revision step of Columbo, as well
as the contextual information in the prompt of the name expansion step.
So ColumboNG only has access to the column names (like NameGuess), but
it still uses rules, chain-of-thought reasoning, and in-context learning. 

Table \ref{colng_ng} shows the results (only for the EM accuracy, for
space reasons). The table shows that compared to the original Columbo,
ColumboNG reduces the EM accuracy by 1.3-22.9\% (see Column
``EM-diff1''). This is as expected. It suggests that exploiting new
information (e.g., the target table names, the names of other tables,
etc.) does help improve accuracy, in some cases significantly.

But interestingly, even though exploiting the exact same information
as NameGuess, ColumboNG still improves accuracy, by 0.5-21.8\% (see
Column ``EM-diff2''). This suggests that the innovations introduced by
ColumboNG, such as rules, using chain-of-thought reasoning, and
in-context learning, do help improve accuracy, in some cases
significantly.

\begin{figure}[t]
    \centering
    \includegraphics[width=0.99\linewidth]{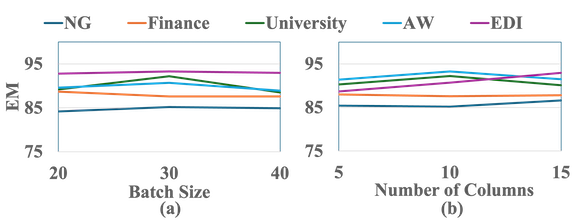}
    \caption{Accuracy of Columbo as we vary the batch size $k$ and the
    number of columns $p$.}
    \label{sens}
\end{figure}

\paragraph{Sensitivity Analysis:} Finally, we examine how Columbo's accuracy
changes as we vary the major parameters. Recall that the Summarizer clusters
the tables in batches (of size $k$). We want to know how the order in
which the tables are processed impacts the accuracy. So we ran Columbo
3 times, using the original default table order, and 2 more orders in
which we randomly shuffled the tables. The results show that the EM
accuracy changes minimally across the three runs, by 0.44-3.58\%.
See the full results in Table \ref{fig:exp_sens_order} in Appendix \ref{asens}.

Next, we vary $k$, the number of tables that the Summarizer clusters
in a single batch, from 20 to 40 (the default value for $k$ is
30). Figure \ref{sens}.a shows that the EM accuracy for all five datasets
fluctuates, but in a small range.  Finally, we vary $p$, the number of
columns that the Generator processes in a single batch, from 5 to 15
(the default value is 10). Figure \ref{sens}.b shows that the EM
accuracy fluctuates but again within a small range. We conclude
that Columbo is robust to small changes in the values of the major
parameters.

\paragraph{Columbo in Production:} We briefly describe how the EDI
team has used Columbo in production on \emph{edirepository.org}.  EDI
is an online data lake where research groups in environmental sciences submit
their data, for other groups to use. EDI has 87K data packages
containing 18K tables. To help researchers find desired tables on EDI,
the EDI team has been trying to assign the concepts from ESCO, a large
ontology that covers environmental domains, to the columns of the 18K
tables.

To date, the EDI team has done such assignments manually, and progress
is slow. In April 2025 they enlisted our help, asking us to assign
ESCO concepts to 36K columns. We frame this problem as a string matching
problem, in which given a table $A$ of 2K concept names in ESCO and a
table $B$ of 36K column names, find all pairs $(x\in A, y\in B)$ that
match.

This problem is challenging because the column names are often
abbreviated and cryptic. Applying standard string matching solutions
(including using embedding vectors) to this problem produced very low
accuracy (less than 30\% on a labeled dataset). So we applied
Columbo to expand all 36K column names, before applying standard ML-based
string matching solutions, achieving 83\% accuracy on the labeled
dataset. The EDI team examined and judged the results ``immensely helpful''.
They have now incorporated Columbo as a part of their assignment workflow.
That is, they use Columbo to expand column names of the new tables,
use our ML-based matcher to match these column names with ontology concepts,
then manually examine the results returned by the matcher to confirm,
reject, or modify the assignments.

This experience, while anecdotal, suggests that expanding column
names is critical for downstream tasks, such as assigning ontology
concepts, and that Columbo is already sufficiently accurate to be
useful in production of some real-world applications.

\paragraph{Table Name Expansion:} Finally, while this paper focuses
on column names, we have conducted preliminary
experiments with expanding table names, on 3 datasets: Finance,
University, and EDI. Asking the LLM to expand the table names, given
the abbreviated column names, produces EM accuracy of 91.3, 98.31,
61.7\%, respectively. Surprisingly, giving the LLM the expanded column
names does not notably improve the EM accuracy, achieving 91.3, 96.6,
and 64.5\%, respectively.

We then examined EDI, where the EM accuracy is lowest, and asked the LLM
to expand the table names, given the gold column names. Also
surprisingly, this improves the EM accuracy by only 0.7\%. We found
that the EDI table names often use phrases that are \emph{not} present
in the column names, e.g., for table ``EVRT1980\_TILLER'', token
``EVRT'' does not appear in any column name, and for table
``2006\_JD\_SnowShrub'', token ``JD'' is the name of the data uploader
and does not appear in any column name.

The above result suggests that expanding table names may require
additional information, such as from textual table descriptions. As a result, we
defer this problem to future research. 

\section{Related Work}\label{sec:related_work}\label{related}

\paragraph{Abbreviation Expansion:}
This task has been studied extensively.  Many
earlier approaches formulate it as a classification problem,
i.e., choosing the most likely expansion from a
predefined candidate set based on surrounding text
\cite{roark2014hippocratic,gorman-etal-2021-structured-abbreviation,
  ice-tea, pouran2020}.  Other lines of research 
expand abbreviations in specific domains like 
informal text~\cite{gorman-etal-2021-structured-abbreviation} and SMS
messages~\cite{sms-expansion}.  \cite{du2019language}
expand prefix-abbreviations in biomedical text.  These works
differ from ours as they often rely on different types of
context (e.g., free-form text) or target different abbreviation styles
than those found in tables.

The work most closely related to ours is NameGuess~\cite{nameguess},
which specifically expand abbreviated column names in tables.
NameGuess introduced a benchmark dataset (based on synthetic
abbreviations) and showed the potential of LLMs, even outperforming
fine-tuned models with one-shot prompting.  However, NameGuess relies
on the column names themselves, without incorporating table names or
broader schema context, and their evaluation was limited to a single
public dataset.  Another relevant study~\cite{snails} also
investigates abbreviated column names but focuses on identifying the
level of abbreviation and analyzing its impact on downstream tasks
like natural language queries, rather than proposing an expansion
method.  Our work leverages richer schema context and more
sophisticated LLM prompting techniques for improved expansion
accuracy. Finally, the work \cite{anonymous2025realistic} follows up
on \cite{nameguess} and focuses on generating more realistic
abbreviations from English phrases.

\paragraph{Table Understanding and Enrichment:}
Expanding abbreviated column names is a specific instance of the
broader goal of enriching table metadata to enhance data
understanding~\cite{fang2024large}, discovery~\cite{freire2025large},
and usability for downstream tasks.
Table-to-text~\cite{zhao2023investigating, zhao2023qtsumm, tabgenie,
  tableformer, gong19table} and table question
answering~\cite{pal-etal-2023-multitabqa, xie-etal-2022-unifiedskg,
  herzig-etal-2020-tapas} aim at developing models able to understand
structured tabular data and natural language questions to perform
reasoning and tasks across tables.

A significant body of work~\cite{turl22, deuer2024ca, sherlock19,
  duduo21,sato20, autotag21, adatyper23} focuses on inferring the
semantic type of data within table columns (e.g., tagging columns as
'zip code', 'address', 'date').  While related, semantic type
detection differs fundamentally from our task; it is typically framed
as a classification problem (assigning a type from a predefined
ontology) based on column values, whereas we focus on generating a
natural language expansion based on the abbreviated column name and
schema context.

LLMs have also been employed to generate natural
language descriptions for tables \cite{tablegpt, zhang2025autoddg,
  gao2025automatic, tabmeta, han2025leveraging} or individual columns
\cite{wretblad2024synthetic}.  These descriptions provide valuable
semantic context but do not directly address the problem of resolving
cryptic abbreviations within column names themselves.

Other related tasks involve matching schemas across different tables
or mapping table columns to concepts in external knowledge graphs or
ontologies, often leveraging LLMs for their semantic understanding
capabilities~\cite{semtab, yang2025matching,
  vandemoortele2024scalable}.

\section{Conclusion}

Expanding the abbreviated column names for tabular data is critical
for many downstream NLP tasks. In this paper we have
significantly advanced the state of the art for this problem. First,
we showed that synthetic public data used by prior work is not
sufficient for experiments, and we introduced four new datasets in
enterprise/science domains, with real-world abbreviations. Second, we
showed that accuracy measures used by prior work undercount correct
expansions, and we proposed new synonym-aware measures that capture
accuracy much more accurately.  Finally, we developed Columbo, a
powerful LLM-based solution, and described extensive experiments,
which show that Columbo outperforms prior work by 4-29\% accuracy on five
datasets.

For future work we will explore improving Columbo, exploiting data tuples
where available, using in-house LLMs, and developing solutions to expand
abbreviated table names.

\paragraph{Acknowledgments:} We thank Suresh Bathini and Guarav Pathak for their assistance with this paper;
Paul Hanson, Colin Smith, Corinna Gries, Mark Tervo, and Minh Phan for
helping us with EDI; and the anonymous reviewers for their insightful
comments. This paper is supported by NSF grant IIS-2504787 and a grant
from Google Inc.

\section{Limitations}

A limitation of our work is that so far we only consider using table
names and column names as the input to expand column names.  In
real-world datasets, tables could contain additional metadata and data
tuples may be available that could provide useful context.
Furthermore, some datasets include taxonomic information, which could
enhance our summarization and local clustering process.  We focus on
table and column names because they are typically the most essential
and consistently available metadata across datasets. Moreover,
domain-specific abbreviations are also common and can aid in column
name expansion. A potential improvement is to
incorporate domain knowledge into the prompting process to enable more
accurate expansion of such abbreviations.

\bibliography{columbo, termmatcher}
\appendix
\section{Appendix}\label{sec:appendix}\label{append}

\subsection{Sample Column Names \& Gold Expansions}\label{sample-cols}

\begin{figure*}[t!]
    \centering
    \begin{subfigure}[t]{0.32\textwidth}
      \includegraphics[width=\textwidth]{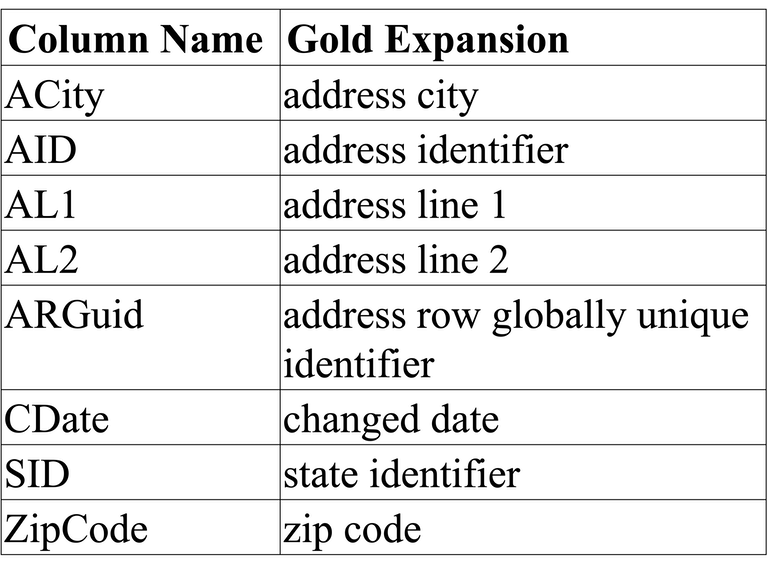}
      \caption{Table name: Address}
    \end{subfigure}
    \hfill
    \begin{subfigure}[t]{0.3\textwidth}
      \includegraphics[width=\textwidth]{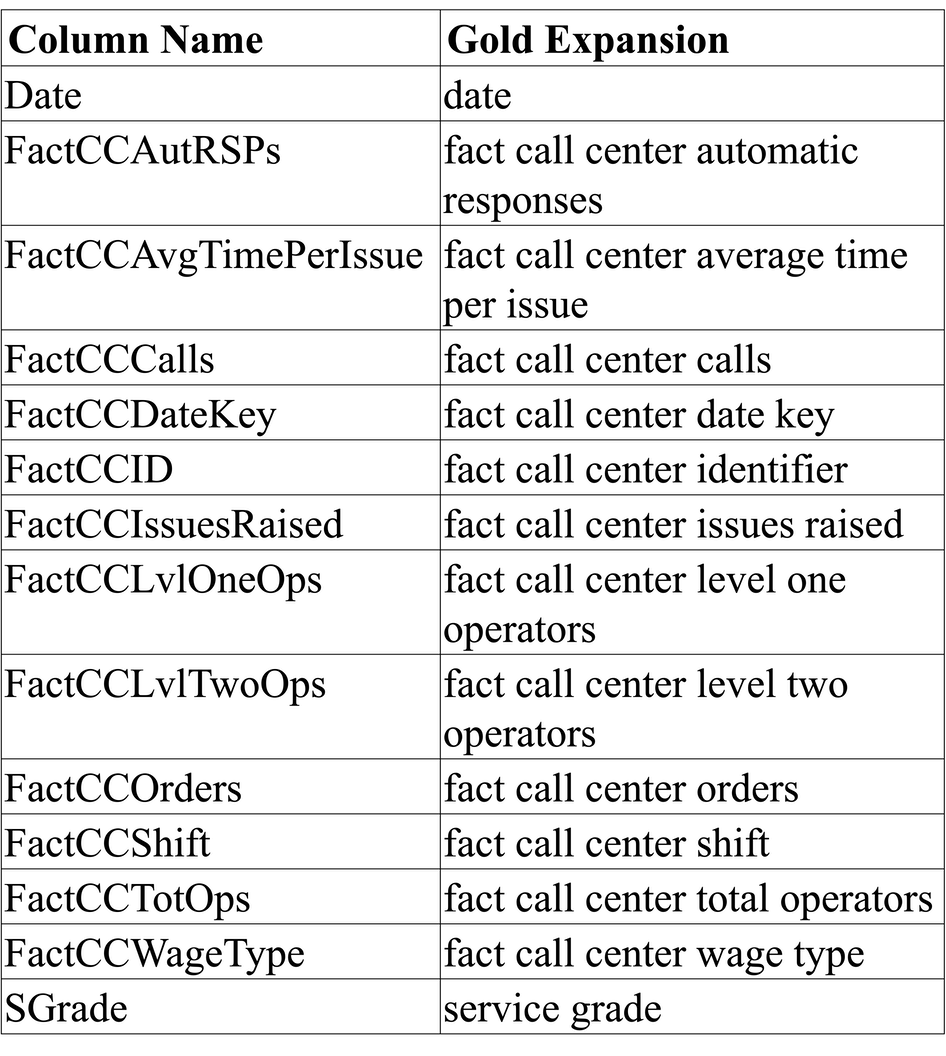}
      \caption{Table name: FactCallCenter}
    \end{subfigure}
    \hfill
    \begin{subfigure}[t]{0.3\textwidth}
      \includegraphics[width=\textwidth]{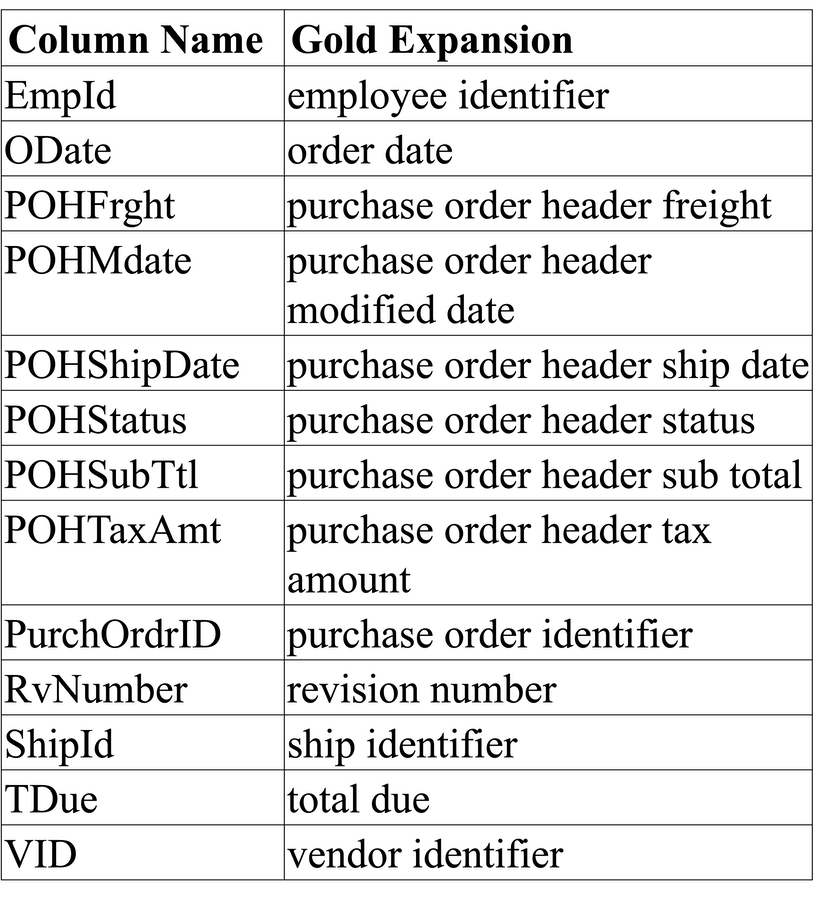}
      \caption{Table name: PurchaseOrderHeader}
    \end{subfigure}

    \caption{Sample column names and gold expansions for the AW dataset.}
    \label{fig:aw_schema}
\end{figure*}

\begin{figure*}[t!]
    \centering
    \begin{subfigure}[t]{0.32\textwidth}
      \includegraphics[width=\textwidth]{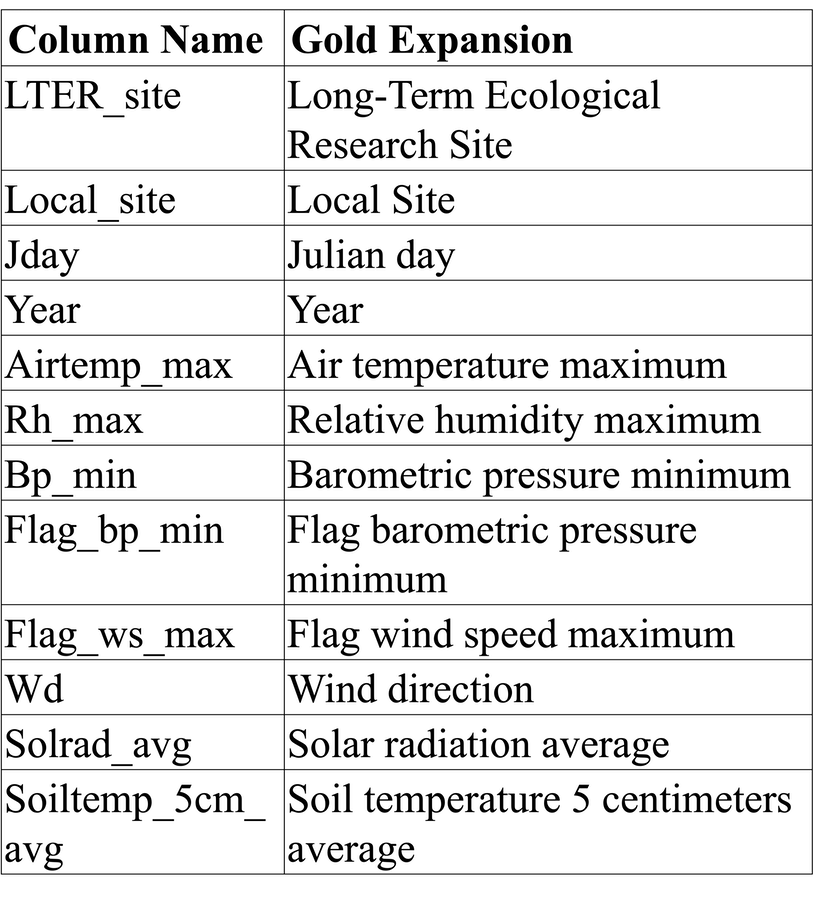}
      \caption{Table name: d-1cr23x-cr1000.daily.ml.data}
    \end{subfigure}
    \hfill
    \begin{subfigure}[t]{0.29\textwidth}
      \includegraphics[width=\textwidth]{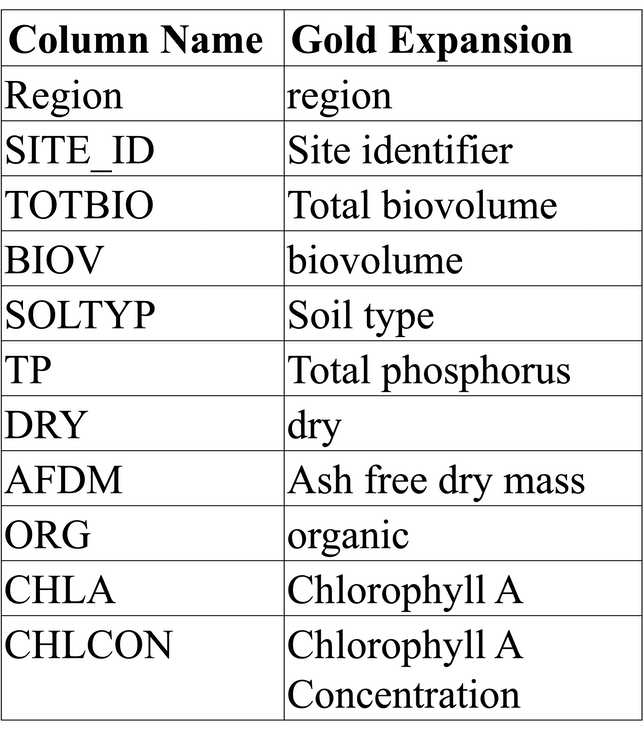}
      \caption{Table name: ST\_PP\_Lahee\_002}
    \end{subfigure}
    \hfill
    \begin{subfigure}[t]{0.27\textwidth}
      \includegraphics[width=\textwidth]{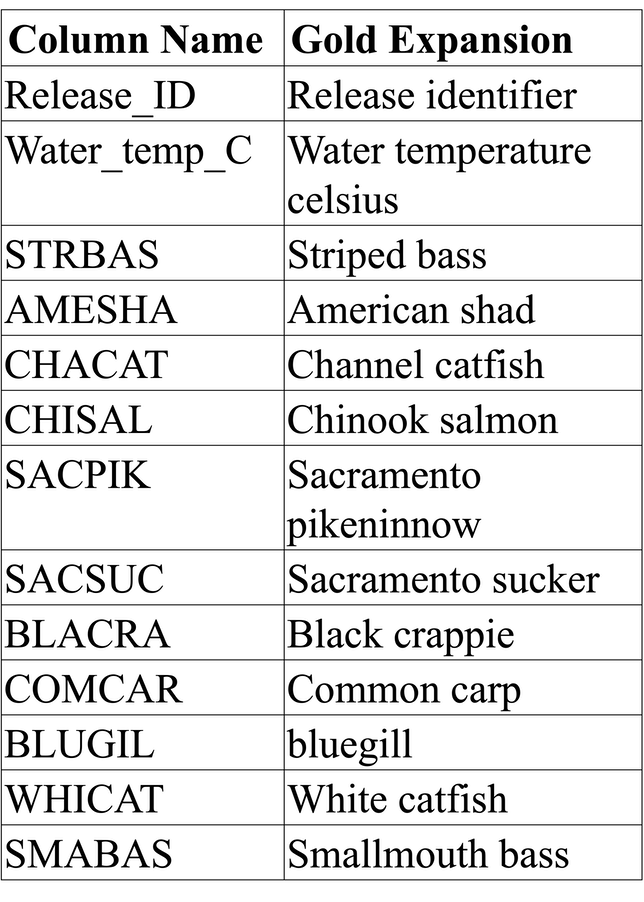}
      \caption{Table name: EDI\_ASB\_SampleDay\_122723}
    \end{subfigure}

    \caption{Sample column names and gold expansions for the EDI dataset.}
    \label{fig:edi_schema}
\end{figure*}

Figures~\ref{fig:aw_schema}-\ref{fig:edi_schema} show sample column names
and gold expansions for the datasets AW and EDI, respectively. 

\subsection{Creating the Datasets}\label{adatasets}

We now describe how the datasets for our experiments were created. 

\paragraph{NameGuess (NG):}
This dataset is provided by~\cite{nameguess}. It comes from the
tables provided by government data portals (San Francisco, Los
Angeles, and Chicago).  Upon using the original gold expansions
provided by the authors, we manually curated some of the gold
expansions where the original gold expansion did not match the
abbreviated column names.  For example, if the original abbreviated
column is `b\_stop', but the gold expansion is `bus stop only', we
will remove `only' and use `bus stop' as the gold expansion.

\paragraph{Finance \& University:} 
They are two proprietary customer datasets from companies in the
finance and university domains, they natively include table names,
abbreviated column names, and detailed column descriptions, which we
used to manually derive the gold expansions.

\paragraph{Environmental Data Initiative (EDI):}
This dataset comes from the~\cite{edi} platform, a repository rich in
ecological data, containing over 18,000 tables from 27 US ecological
sites.  We randomly sampled $251$ tables from the EDI platform, and
used the extensive provided metadata (project details, table/column
descriptions) to manually create gold expansions for the originally
abbreviated column names.

\paragraph{Adventure Works (AW):} 
This dataset is based on the Adventure Works sample database,
representative of common enterprise schemas. Its original schema
contains full-form names. But in a separate project the participants
have manually abbreviated the column names using the table context.
We manually reviewed these human-generated abbreviations and verified
or corrected gold expansions.

Across all datasets, there were instances where determining a
definitive gold expansion with high confidence was impossible due to
ambiguity or insufficient metadata.  For example, in the EDI dataset,
there are column names `x' and `y' and no definition is provided. In
such cases we cannot determine the gold expansion for these columns.
As another example, in the EDI dataset, in the
`d-1cr23x-cr1000.daily.ml.data' table, there is a column named
`airtemp\_hmp1\_max', however there is no definition of the token
`hmp'  and we
cannot find other information about the token `hmp' in other entries
of the project metadata.  Moreover, LLM may generate various possible
expansions for this token and we cannot determine which one is
correct as we are not domain experts. 

Such columns were excluded from the calculation of evaluation metrics.
However, they were retained as part of the input schemas provided to
the models, as they still contribute valuable contextual information
about the table's structure and domain.

\subsection{Sample LLM Prompt and Output for the Summarizer}\label{sec:prompt_summarizer}\label{asum}

\begin{figure*}[t]
    \centering
    \includegraphics[width=1\linewidth]{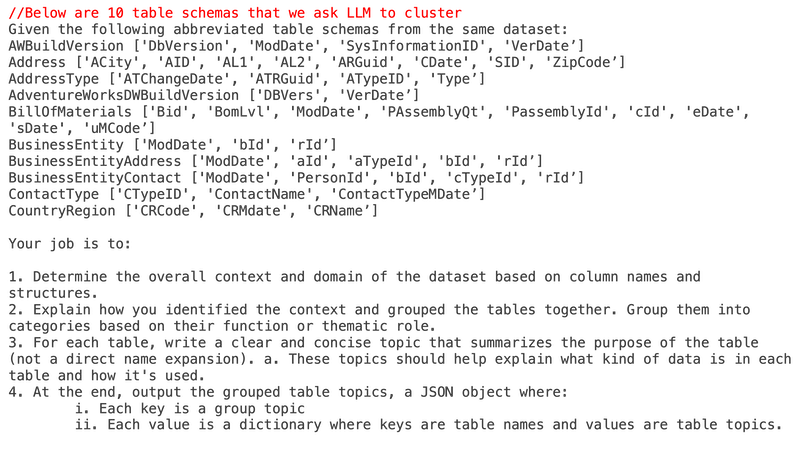}
    \caption{A sample prompt to the LLM to ask it to cluster 10 tables from the AW dataset. Similar prompts
      are used by the Summarizer (but sending 30 tables to the LLM).}
    \label{fig:prompt_clustering}
\end{figure*}

\begin{figure*}[t]
    \centering
    \includegraphics[width=0.75\linewidth]{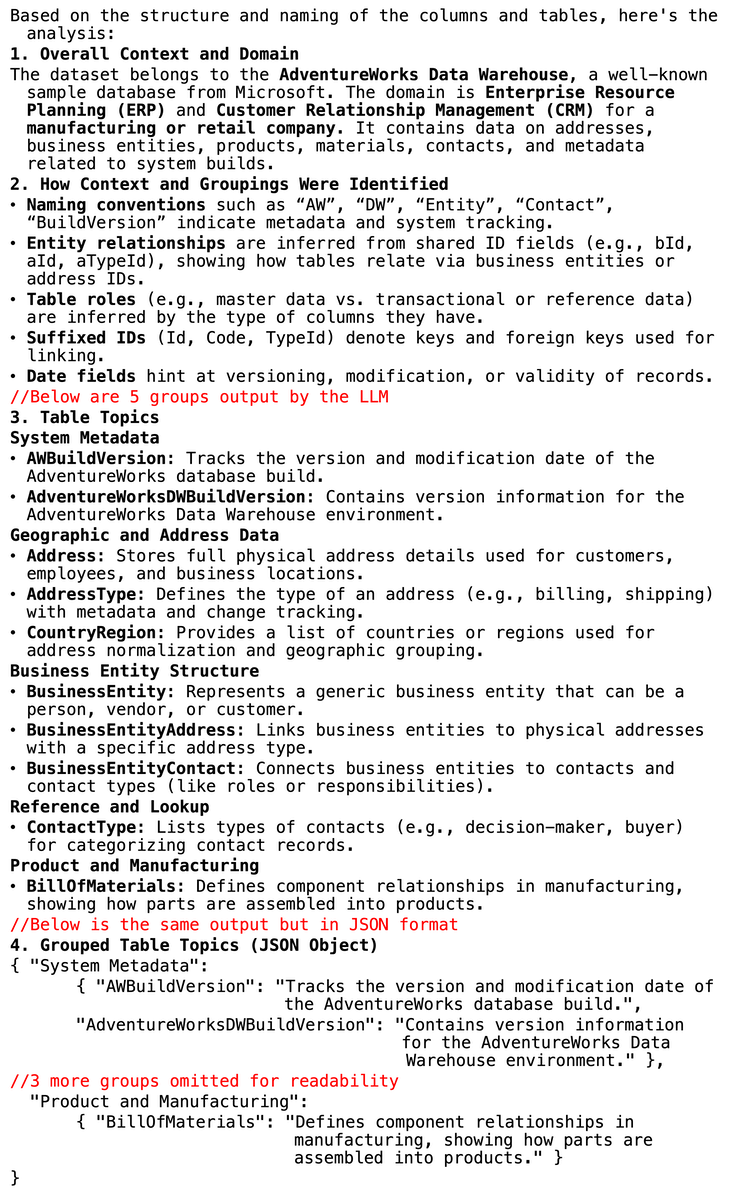}
    \caption{A sample output by GTP-4o, in response to the prompt in Figure \ref{fig:prompt_clustering}.
      Note how the LLM groups the 10 input tables into 5 groups, with 2, 3, 3, 1, 1 tables in each group,
      respectively. The first group has the summary ``System Metadata'' and has 2 tables. The second group
      has the summary ``Geographic and Address Data'' and has 3 tables, and so on. Note also that the LLM
      produces for each table a short summary.}
    \label{fig:output_clustering}
\end{figure*}

Figure~\ref{fig:prompt_clustering} shows a sample prompt to the LLM to ask it to cluster
10 given tables. Figure~\ref{fig:output_clustering} shows a sample output from GPT-4o
for the above prompt.

\subsection{Sample LLM Prompt and Output for the Generator}\label{sec:prompt_generator}\label{agen}
\begin{figure*}[t]
    \centering
    \includegraphics[width=0.75\linewidth]{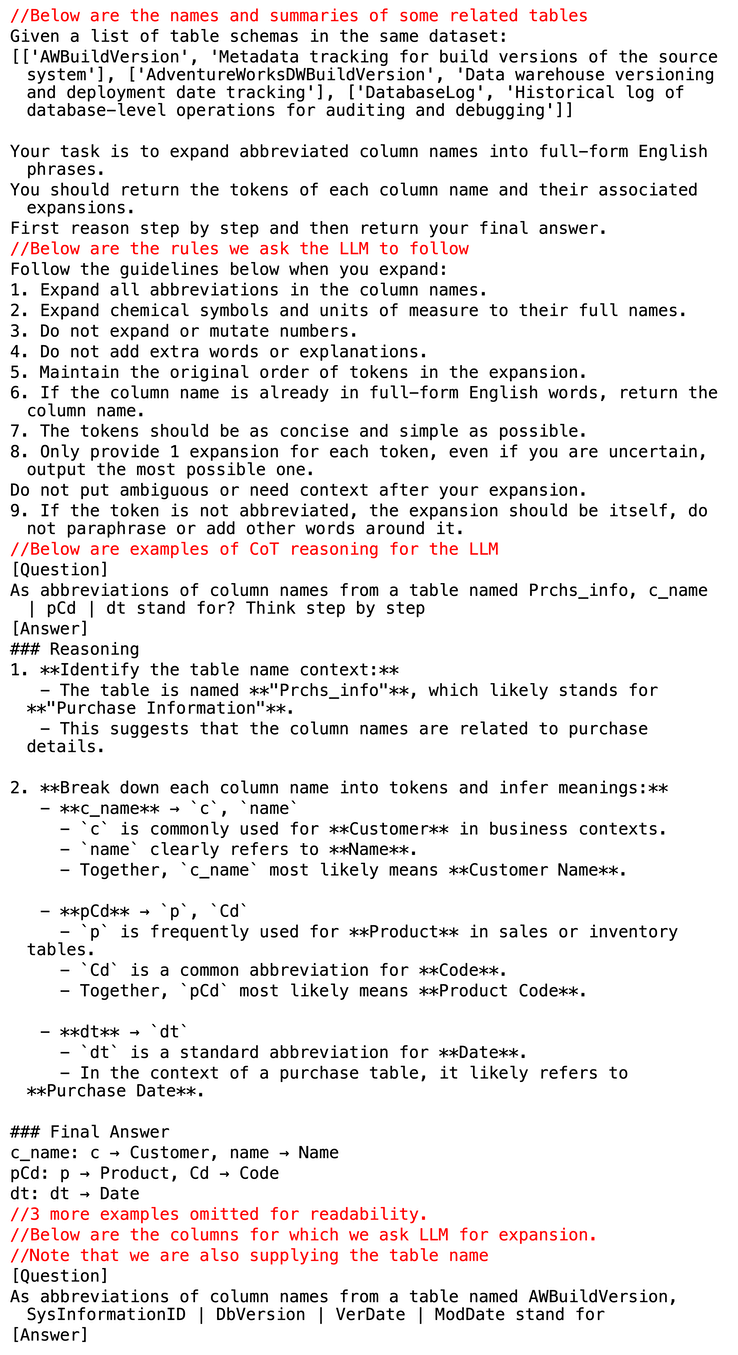}
    \caption{A sample prompt to the LLM to ask it to expand 4 column names of a table. Similar prompts
      are used by the Generator, but asking to expand up to 10 column names. Note how we provide
      the names and summaries of some related tables, the rules, examples of CoT reasoning. Then
      we finally provide the 4 abbreviated column names (and the table name) in the last three lines
      of the prompt, and ask the LLM to expand those column names.}
    \label{fig:prompt_expansion}
\end{figure*}

\begin{figure*}[t]
    \centering
    \includegraphics[width=0.75\linewidth]{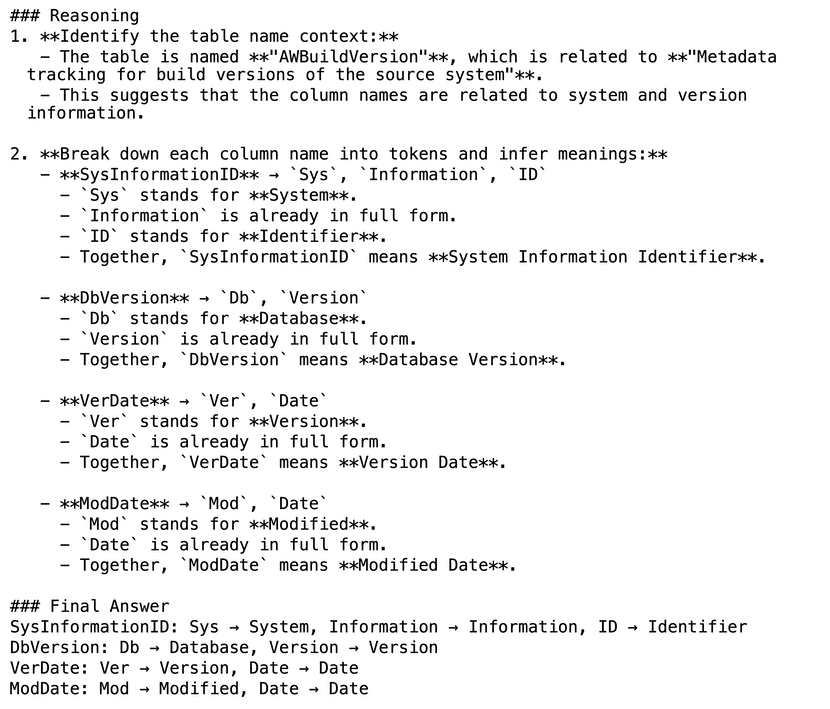}
    \caption{A sample output from GPT-4o that provides the E2 records for the 4 column names mentioned
      in the prompt of the previous figure.}
    \label{fig:output_expansion}
\end{figure*}

Figure~\ref{fig:prompt_expansion} shows a sample prompt to the LLM to ask it to expand 4 column names from a table.
Similar prompts are used by the Generator (but asking the LLM to expand 10 column names).
Figure~\ref{fig:output_expansion} shows a sample output from the LLM for the above prompt. 

\subsection{Sample LLM Prompt and Output for the Reviser}\label{sec:prompt_reviser}\label{arev}

\begin{figure*}[t]
    \centering
    \includegraphics[width=0.75\linewidth]{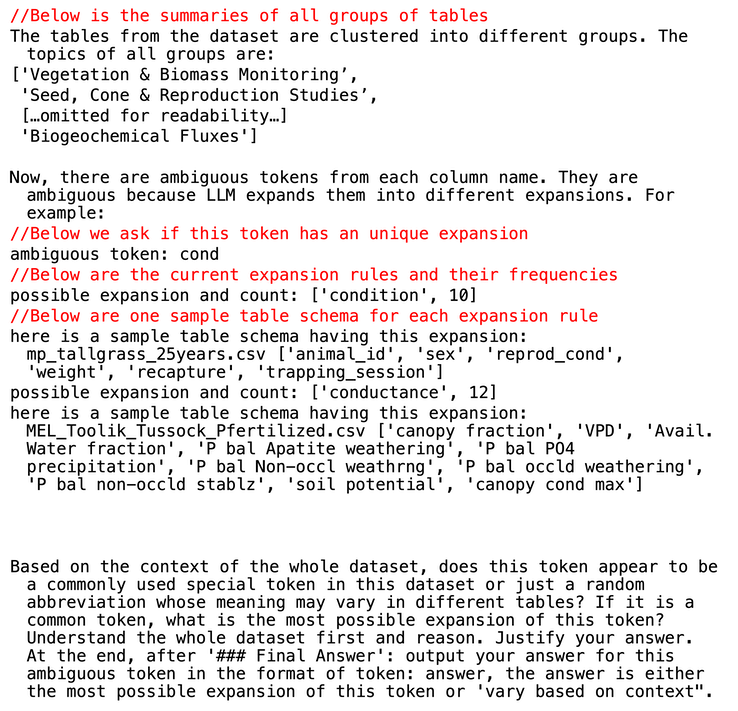}
    \caption{A sample prompt asking the LLM if the token ``cond'' has a unique expansion.} 
    \label{fig:prompt_context_dependency}
\end{figure*}

\begin{figure*}[t]
    \centering
    \includegraphics[width=0.7\linewidth]{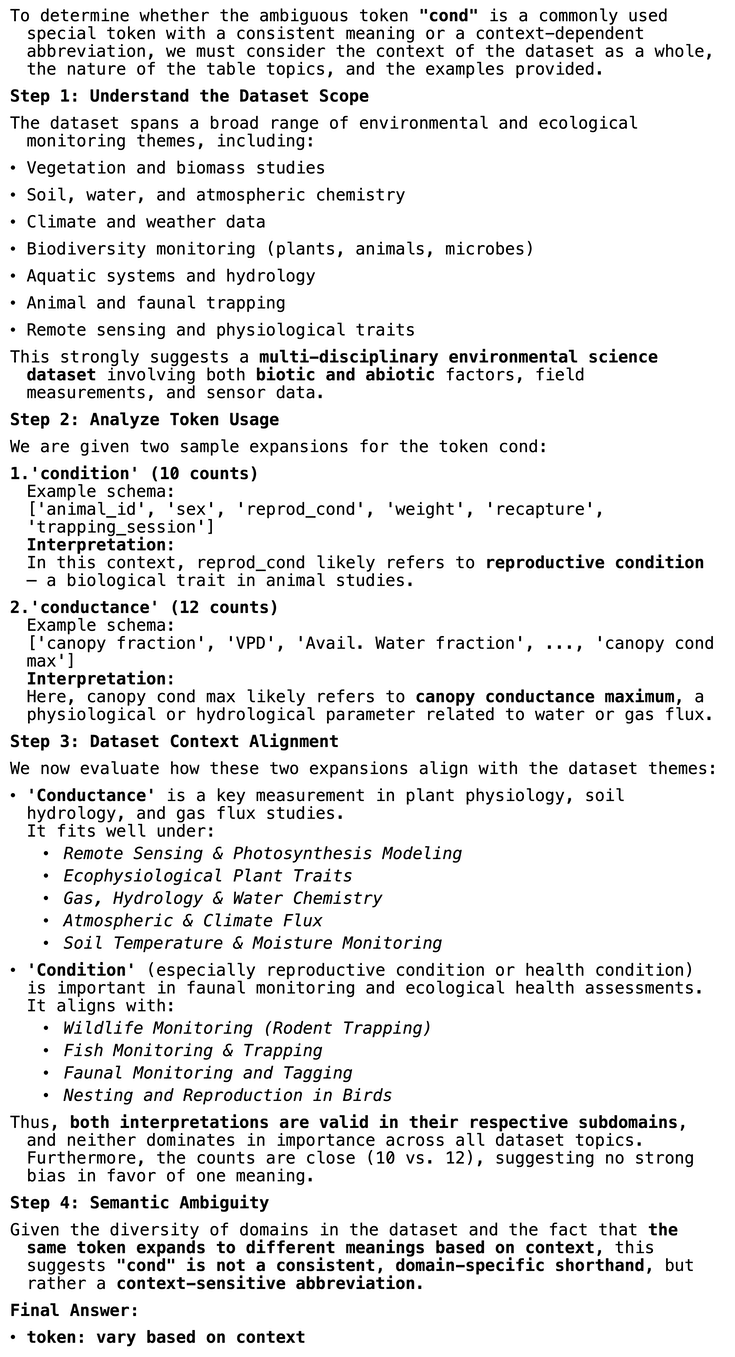}
    \caption{A sample output from GPT-4o stating that token ``cond'' does not have a unique
      expansion rule (see the last line of the sample output.}
    \label{fig:output_revising}
\end{figure*}

Figure~\ref{fig:prompt_context_dependency} shows a sample prompt to the LLM asking if a given token has a unique
expansion. Similar prompts are used by the Reviser. Figure~\ref{fig:output_revising} shows
a sample output by GPT-4o that determines that the given token does not have a unique expansion. 

\subsection{Ablation Studies}\label{aablation}
\begin{figure*}[!t]
  \centering
  
  \begin{subfigure}{0.45\linewidth}
    \includegraphics[width=\linewidth]{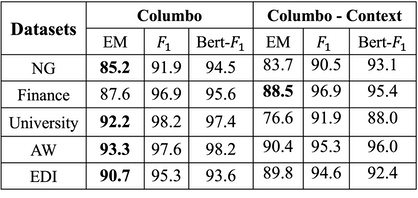}
    \caption{The ablation study where Columbo does not exploit contexts.}
  \end{subfigure}\hfill
  \begin{subfigure}{0.45\linewidth}
    \includegraphics[width=\linewidth]{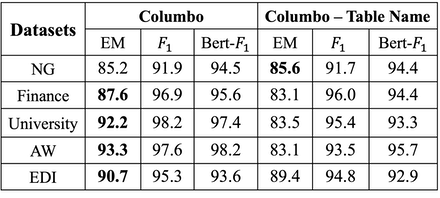}
    \caption{The ablation study where Columbo does not exploit table names.}
  \end{subfigure}

  \begin{subfigure}{0.45\linewidth}
    \includegraphics[width=\linewidth]{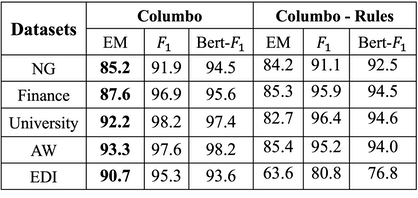}
    \caption{The ablation study where Columbo does not exploit rules.}
  \end{subfigure}\hfill
  \begin{subfigure}{0.45\linewidth}
    \includegraphics[width=\linewidth]{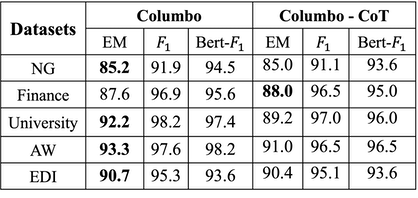}
    \caption{The ablation study where Columbo does not exploit CoT reasoning.}
  \end{subfigure}

  \begin{subfigure}{0.45\linewidth}
    \includegraphics[width=\linewidth]{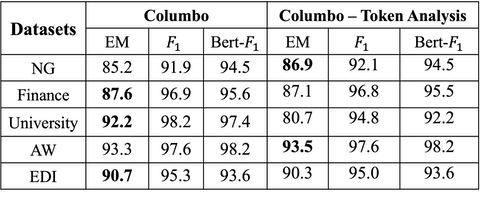}
    \caption{The ablation study where Columbo does not exploit token-level analysis.}
  \end{subfigure}

  \caption{The Ablation studies of Columbo}
  \label{fig:five}
\end{figure*}

Figure \ref{fig:five} shows the full result of the ablation studies,
as we examine the effect of five major components of Columbo. 

\subsection{Sensitivity Analysis}\label{asens}

\begin{figure*}[!t]
    \centering
    \includegraphics[width = 1\linewidth]{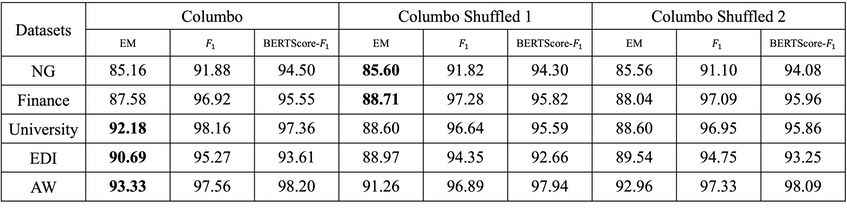}
    \caption{The accuracy of Columbo as we try three different orderings of the tables, for the Summarizer.}
    \label{fig:exp_sens_order}
\end{figure*}

Figure~\ref{fig:exp_sens_order} shows the full result of the sensitivity
analysis with respect to the ordering of tables for the Summarizer.


\end{document}